\documentclass[10pt,twocolumn,letterpaper]{article}

\usepackage{cvpr}              

\usepackage[dvipsnames]{xcolor}

\definecolor{cvprblue}{rgb}{0.21,0.49,0.74}
\usepackage[pagebackref,breaklinks,colorlinks,citecolor=cvprblue]{hyperref}

\usepackage{graphicx}
\usepackage{amsmath}
\usepackage{amssymb}
\usepackage{booktabs}
\usepackage[accsupp]{axessibility}
\usepackage[ruled,vlined]{algorithm2e}
\usepackage{float}
\usepackage{colortbl}
\usepackage{amssymb}
\usepackage{makecell}
\usepackage{tablefootnote}
\usepackage{multirow}
\usepackage{pifont}
\usepackage[font=small]{caption}
\usepackage{subcaption}
\usepackage{paralist}
\usepackage[misc]{ifsym}
\usepackage{arydshln}
\usepackage{bm}

\definecolor{rowblue}{RGB}{198,234,251}   
\definecolor{rowgreen}{RGB}{209,239,191}   
\definecolor{rowgray}{RGB}{245,245,245}   
\definecolor{highgreen}{RGB}{0,139,69}
\definecolor{commentcolor}{RGB}{237,2,140}   

\def\paperID{132} 
\def\confName{CVPR}
\def\confYear{2024}

\title{VideoMAC: Video Masked Autoencoders Meet ConvNets}

\author{Gensheng Pei\thanks{Equal contribution.}, Tao Chen\footnotemark[1], Xiruo Jiang, Huafeng Liu, Zeren Sun\thanks{Corresponding author.}, Yazhou Yao\footnotemark[2] \\
\small{School of Computer Science and Engineering, Nanjing University of Science and Technology} \\
\small{\url{https://github.com/NUST-Machine-Intelligence-Laboratory/VideoMAC}}\\
}

\begin{document}
\maketitle
\begin{abstract}
Recently, the advancement of self-supervised learning techniques, like masked autoencoders (MAE), has greatly influenced visual representation learning for images and videos.
Nevertheless, it is worth noting that the predominant approaches in existing masked image / video modeling rely excessively on resource-intensive vision transformers (ViTs) as the feature encoder.
In this paper, we propose a new approach termed as \textbf{VideoMAC}, which combines video masked autoencoders with resource-friendly ConvNets.
Specifically, VideoMAC employs symmetric masking on randomly sampled pairs of video frames.
To prevent the issue of mask pattern dissipation, we utilize ConvNets which are implemented with sparse convolutional operators as encoders. 
Simultaneously, we present a simple yet effective masked video modeling (MVM) approach, a dual encoder architecture comprising an online encoder and an exponential moving average target encoder, aimed to facilitate inter-frame reconstruction consistency in videos.
Additionally, we demonstrate that VideoMAC, empowering classical (ResNet) / modern (ConvNeXt) convolutional encoders to harness the benefits of MVM, outperforms ViT-based approaches on downstream tasks, including video object segmentation (+\textbf{5.2\%} / \textbf{6.4\%} $\mathcal{J}\&\mathcal{F}$), body part propagation (+\textbf{6.3\%} / \textbf{3.1\%} mIoU), and human pose tracking (+\textbf{10.2\%} / \textbf{11.1\%} PCK@0.1).
\end{abstract}

\begin{figure}
    \begin{center}
        \includegraphics[width=1.0\linewidth]{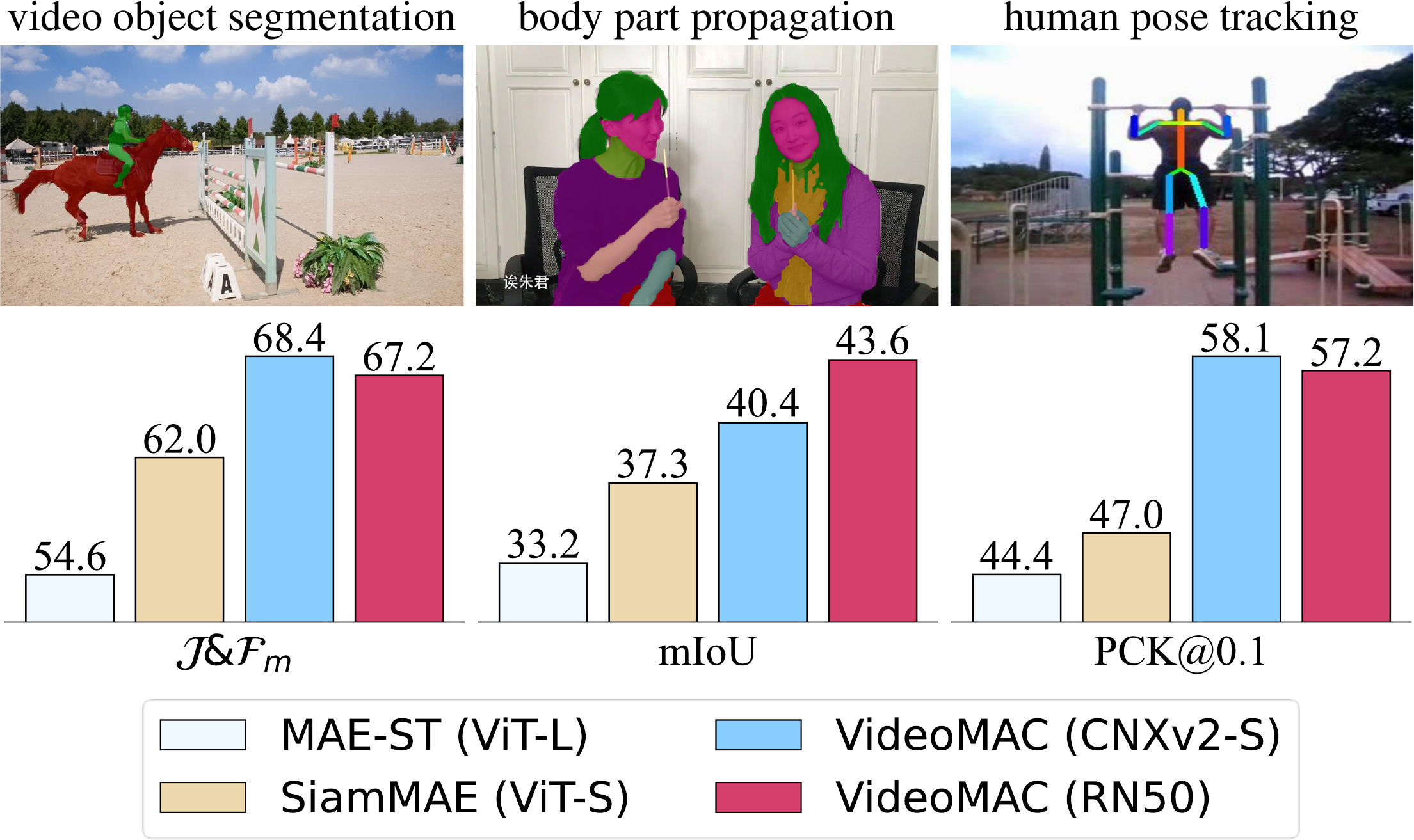}
    \end{center}
    \vspace{-0.6cm}
    \caption{State-of-the-art MAE architectures (\eg, MAE-ST~\cite{mae_st} and SiamMAE~\cite{gupta2023siamese}) in masked video modeling commonly employ ViT-based encoders.
    We propose \textbf{VideoMAC}, a new video masked autoencoder built using pure ConvNets.
    In this study, we model VideoMAC with two of the most emblematic families of ConvNets, namely ConvNeXtv2~\cite{woo2023convnext} and ResNet~\cite{he2016deep}.
    Notably, VideoMAC exhibits superior performance on a range of downstream tasks, \eg, video object segmentation on DAVIS17~\cite{davis17}, body part propagation on VIP~\cite{zhou2018adaptive}, and human pose tracking on JHMDB~\cite{jhuang2013towards}, compared to ViT-based methods~\cite{mae_st,gupta2023siamese}.}\label{fig1}
    \vspace{-0.6cm}
\end{figure}

\section{Introduction}
Spurred by the success of masked language modeling in natural language processing~\cite{BERT,clark2020electra,brown2020language}, transformer-based visual models, notably vision transformers (ViT)~\cite{dosovitskiy2021vit} adopt masked image modeling (MIM) are attracting attention~\cite{beit,he2022mae,SimMIM,assran2022msn,fang2023eva,liu2023improving}. This strategy involves masking or corrupting a portion of information within an image and then requiring the model to recover the masked portion, encouraging the model to learn useful image representations.

With the successful application of MIM, some pioneering works~\cite{mae_st,tong2022videomae} have shifted the focus of video representation learning to masked video modeling (MVM), yielding promising results in downstream video tasks. However, features learned by MVM are designed explicitly for pixel reconstruction tasks. They perform well when fine-tuned for downstream video tasks but exhibit subpar performance without specific fine-tuning~\cite{gupta2023siamese}. Their performance is even inferior to MIM methods~\cite{he2022mae,lee2023exploring}, and it lags significantly behind supervised methods~\cite{he2016deep,liu2022convnet}.

We strive to discern the factors contributing to this issue. To the best of our knowledge, most current MVM methods are based on the isotropic design of ViT~\cite{mae_st,tong2022videomae,wu2023dropmae,gupta2023siamese}. The reason is that early MIM / MVM methods are explicitly developed for ViT, as they allow the use of only visible patches as the sole input to the encoder, which aligns well with the ViT architecture. For instance, the representative approach, masked autoencoders (MAE)~\cite{he2022mae}, enhances the efficiency and effectiveness of model pre-training by focusing predominantly on the unmasked regions.
However, previous MVM approaches~\cite{mae_st,tong2022videomae}, confined by the plain vision transformer design devoid of a hierarchical structure tend to result in inadequate spatial modeling capabilities for local features~\cite{liu2021swin,wang2021pyramid,ryali2023hiera,zhang2023hivit,gao2022mcmae}. Such capabilities are crucial for dense downstream tasks like segmentation and detection. Moreover, the uniform reliance on ViT in existing MVM methods has resulted in significant computational resource consumption~\cite{mae_st,tong2022videomae,wang2023videomae}.
One might first consider ConvNets~\cite{lecun1998lenet5,krizhevsky2012alexnet,simonyan2015vgg,he2016deep,tan2019efficientnet,liu2022convnet}, characterized by their low resource consumption and built-in hierarchical structure, as an ideal solution to tackle the aforementioned issues.
Nevertheless, ConvNets appears understudied in MVM.
Consequently, it raises the question: \textit{can ConvNets, endowed with a pyramid vision architecture, guide MVM towards the era of hierarchical pre-training?}

\begin{figure}
    \begin{center}
        \includegraphics[width=0.93\linewidth]{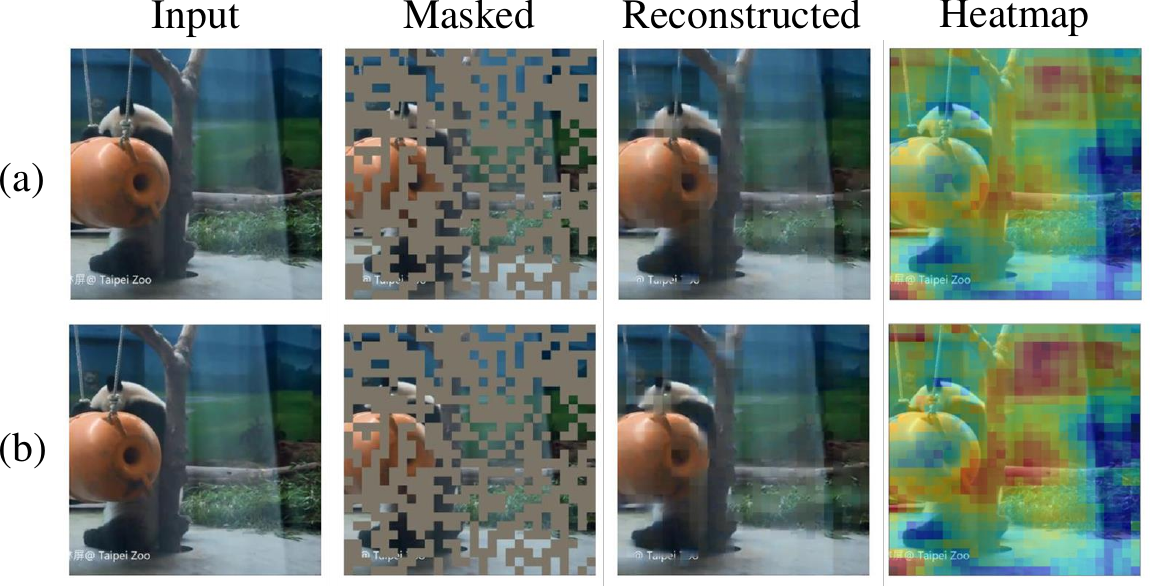}
    \end{center}
    \vspace{-0.6cm}
    \caption{Visualization of the heatmap during the reconstruction of masked patches in a random frame pair. It is evident that our approach highlights similar regions for both (a) past and (b) current frames, proficiently reconstructing colors and contours.}\label{fig:rec_heatmap}
    \vspace{-0.6cm}
\end{figure}

To address this concern, we propose a simple yet potent MVM approach, referred to as \textit{\textbf{Video} \textbf{M}asked \textbf{A}utoencoders with \textbf{C}onvNets (VideoMAC)}.
Uniquely, our method represents an initial endeavor to supplement video pre-training by integrating hierarchical ConvNets.
However, applying ConvNets to MVM proves more complex than utilizing ViTs. Typical ConvNets might be incompatible with MVM due to the employment of dense sliding windows with overlap, resulting in information leakage within masked regions.
To efficiently incorporate mask modeling into ConvNets, we opt for sparse convolution over dense convolution to restructure the original ConvNets, an approach validated in recent studies~\cite{woo2023convnext,tian2023designing} to effectively maintain the structure of the masked region when utilizing ConvNets. 
Another hurdle for MVM is the effective incorporation of temporal data. Current MVM pre-training methods~\cite{mae_st,tong2022videomae} involving multi-frame inputs are primarily used in downstream resource-intensive tasks such as video action recognition. 
Although methods using two-frame inputs for pre-training lower resource usage~\cite{wu2023dropmae,gupta2023siamese}, they typically employ the Siamese architecture, necessitating robust skills for designing reconstruction targets. Importantly, all aforementioned methods utilize computationally intensive, non-hierarchical ViT~\cite{dosovitskiy2021vit} as their encoders.
To this end, we utilize pairs of frames as inputs and introduce the reconstruction consistency loss, a streamlined extension of MAE to accommodate spatio-temporal data.
We formulate online networks through gradient optimization and establish target networks by updating parameters via an exponential moving average (EMA).
This allows for individual recovery of masked frame pairs and computation of the reconstruction consistency loss between them. As illustrated in \cref{fig:rec_heatmap}, our approach can focus on similar regions between two frames and reconstruct the masked portions efficiently.

Notably, our approach transcends previous state-of-the-art techniques in three downstream tasks, \eg, video object segmentation on DAVIS17~\cite{davis17}, body part propagation on VIP~\cite{zhou2018adaptive}, and human pose tracking on JHMDB~\cite{jhuang2013towards}. Moreover, we present promising results utilizing VideoMAC in fine-tuning for image recognition on ImageNet1K~\cite{deng2009imagenet}. To summarize, our main contributions of this work include:
\begin{compactitem}
  \item We propose a simple yet potent solution of masked video autoencoders, marking the initial endeavor to successfully implement MVM using pure ConvNets.
  \item We introduce the reconstruction consistency loss to enable elegant modeling of spatio-temporal data.
  \item Our approach significantly outperforms existing ViT-based MVM methods in various downstream tasks.
\end{compactitem}

While the focus of the computer vision community is increasingly shifting towards vision transformers, we trust that VideoMAC will rekindle interest in exploring ConvNet-based MVM, thereby catalyzing the discovery of new potentials across a variety of video tasks.

\section{Related Work}
\noindent\textbf{Masked Image Modeling.}
Self-supervised learning (SSL) has been rapidly developing in computer vision, demonstrating substantial potential for deriving valuable representations from extensive, unlabeled datasets.
A key advantage of SSL~\cite{he2020moco,chen2020simclr,grill2020bootstrap,chen2021simsiam,caron2021emerging,caron2020swav,clip,beit,touvron2021deit,zhou2022ibot,yao2021non} is its independence from costly human annotations, thereby providing robust prior knowledge across various visual tasks. The success of masked language modeling in NLP~\cite{BERT,clark2020electra,dong2019unified,brown2020language,radford2019language} has catalyzed a rising interest in masked image modeling (MIM) for visual pre-training.
This strategy involves masking or corrupting a portion of information within an image and then requiring the model to recover the masked portion, encouraging the model to learn useful image representations~\cite{beit,he2022mae,SimMIM}. Notably, pioneering works~\cite{woo2023convnext,tian2023designing} have recently applied ConvNets to MIM pre-training, paving the way for masking modeling methods based on ConvNets.

\noindent\textbf{Masked Video Modeling.}
Recently, researchers have embarked on exploring MVM approaches~\cite{mae_st,wu2023scalable,wu2023dropmae,gupta2023maskvit} to comprehend unsupervised video representations, exhibiting effectiveness across a multitude of downstream tasks~\cite{caelles2017one,zhu2017flow,wang2019learning,wang2019zero,jabri2020space,lu2020learning,lu2020video,pei2022hfan,li2022locality,zhang2023boosting,li2023unified,pei2023hgpu,pei2023hcpn,zhou2023survey}.
However, current MVM methodologies~\cite{tong2022videomae,wang2023videomae,gupta2023siamese,wang2023masked} primarily depend on the isotropic ViT, a visual backbone compatible with masking modeling.
Furthermore, these ViT-based MVM strategies may encounter challenges with inadequate local feature modeling when dealing with dense tasks such as segmentation and detection due to their non-hierarchical structure~\cite{zhang2023hivit,liu2022video,liu2021swin,wang2021pyramid}.
To the best of our knowledge, no MVM approaches are explicitly designed for hierarchical network-based architectures, especially ConvNets.

It is worth noting that while the isotropic ViT is well suited for the implementation of MVM methods~\cite{mae_st,tong2022videomae,gupta2023maskvit}, this is exceptionally difficult for ConvNets.
In this paper, our VideoMAC is committed to developing an efficient video pre-training tool that can equip ConvNets with superior, or at least comparable, performance to ViTs.

\section{Approach}
Our VideoMAC builds upon MAE~\cite{he2022mae} and, pioneeringly, facilitates masked video modeling with a purely convolutional architecture.
As depicted in Fig.~\ref{fig:pipeline}, VideoMAC is designed to reconstruct masked patches, which are randomly and symmetrically masked on frame pairs at a high masking ratio (0.75 as used in this study).
The framework achieves such reconstructions by exclusively delivering visible patches to both the online and target encoders. In this section, we describe its main components in detail.

\subsection{Preliminaries}
Recently, masked autoencoders (MAE)~\cite{he2022mae} introduces a novel methodology to the domain of self-supervised visual representation learning.
The primary process involves training a model to reconstruct masked areas in images whose contents are partially occluded.
The fundamental architecture of a standard MAE model is composed of three core elements: the masking design, an encoder, and a decoder.
In mathematical terms, given an image $\bm{I}\in\real^{3 \times H \times W}$, the MAE model patchifies $\bm{I}$ into $N$ distinct, non-overlapping patches, represented by $\hat{\bm{I}}\in\real^{N \times (P^2 \times 3)}$, each with a size of $P$.
Masks are randomly applied to a subset of these patches, denoted by the subset $\mathcal{M}$.

The encoder $f_{\mathcal{E}}$ processes the visible patches $\mathcal{V}$, producing latent representations that are represented as $z=f_{\mathcal{E}}(\mathcal{V})$.
Then, the decoder $f_{\mathcal{D}}$ attempts to restore the masked subset $\mathcal{M}$ utilizing these latent representations, generating $\bm{O}=f_{\mathcal{D}}(z;\mathcal{M})$, where $\bm{O}\in\real^{N \times (P^2 \times 3)}$ depicts the reconstructed patches.
The MAE model utilizes the mean-squared error (MSE) to compute the reconstruction loss $\mathcal{L}_{rec}$, which is solely calculated based on the masked patches and is formulated as $\mathcal{L}_{rec}=\frac{1}{|\mathcal{M}|}\sum_{i\in\mathcal{M}}||\hat{\bm{I}}_{i}-\bm{O}_{i}||_2^2.$

\subsection{Video Masked Autoencoder with ConvNets}
\noindent\textbf{Masking.}
In this work, we randomly select two consecutive frames, $\bm{I}^{(1)}, \bm{I}^{(2)}\in\real^{3 \times H \times W}$, and divide them into $N$ non-overlapping patches, represented as $\hat{\bm{I}}^{(1)}, \hat{\bm{I}}^{(2)}\in\real^{N \times (P^2 \times 3)}$.
Regarding the mask configuration, our approach adopts symmetric masking (\cref{tab:asym_mask}) to guarantee consistency in the positions of the masked areas between frame pairs.
Given the consecutive frames, $\mathcal{M}^{(1)}$ and $\mathcal{M}^{(2)}$ constitute the subsets of masked patches, whereas $\mathcal{V}^{(1)}$ and $\mathcal{V}^{(2)}$ correspond to the visible patches.

It is inherent that images and videos possess a greater level of information redundancy~\cite{mae_st,he2020moco} compared to languages. Accordingly, we employ a masking ratio of 0.75 (\cref{tab:mask_ratio}) in our methodology. Subsequent empirical analyses (\cref{sec:ablation}) have demonstrated an improvement in performance with higher masking ratios.
Nevertheless, employing excessively high masking ratios may induce the risk of training collapse, thereby leading to sudden and substantial decreases in model performance.

\begin{figure}
    \vspace{-0.1cm}
    \begin{center}
        \includegraphics[width=1.0\linewidth]{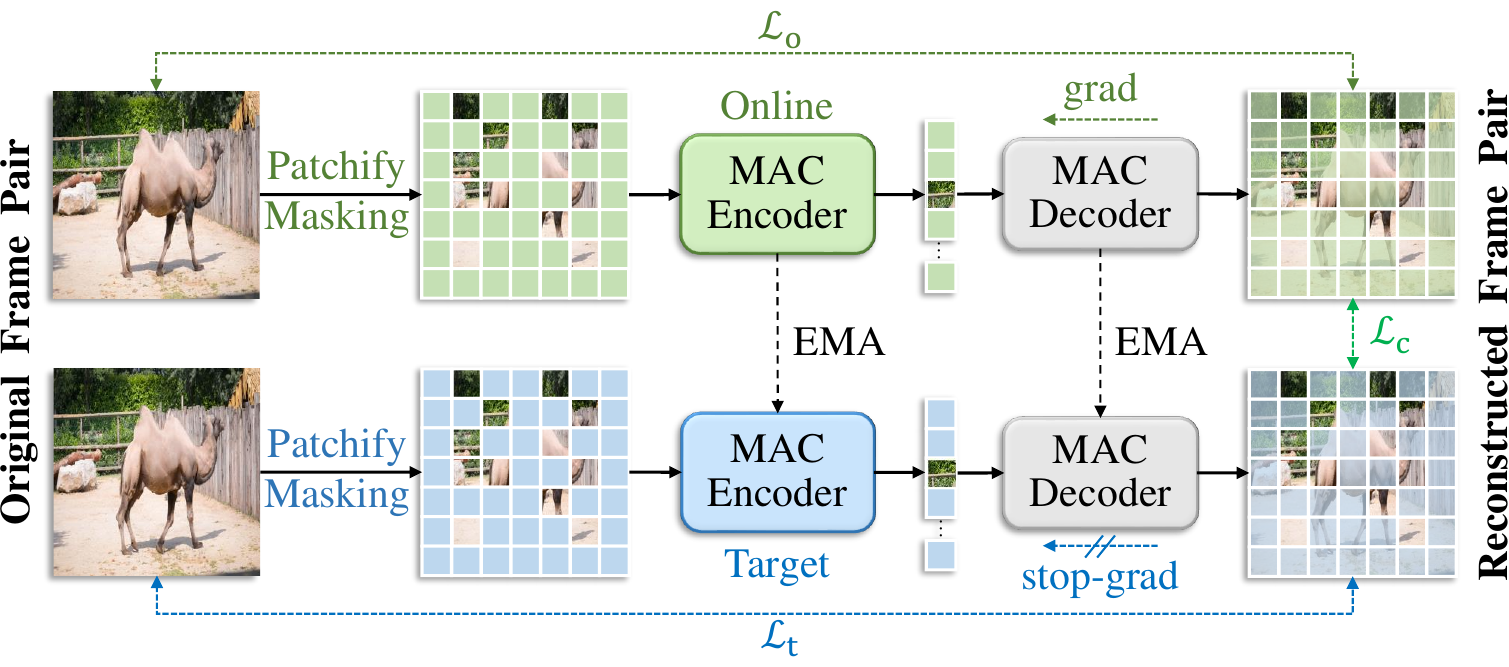}
    \end{center}
    \vspace{-0.65cm}
    \caption{An illustration of VideoMAC for ConvNet-based MVM. During pre-training, we mask 75\% of symmetric patches from two frames randomly. In our VideoMAC, the MVM of frame pairs is achieved by an online network optimized by gradients (\textcolor[RGB]{197,224,180}{$\blacksquare$}, online loss $\mathcal{L}_{o}$) and a target network updated by EMA (\textcolor[RGB]{189,215,238}{$\blacksquare$}, target loss $\mathcal{L}_{t}$). $\mathcal{L}_{c}$ is computed as the reconstruction consistency loss between reconstructed patches of frame pairs.}
    \label{fig:pipeline}
    \vspace{-0.5cm}
\end{figure}

\noindent\textbf{MAC Encoder.}
In ViT-based models, the prevention of information leakage from masked regions is comparatively straightforward~\cite{woo2023convnext,zhang2023hivit}. These models solely employ the visible patches as the exclusive input to the encoder, thereby allowing ViT-based approaches~\cite{mae_st,tong2022videomae,wu2023dropmae,gupta2023siamese} to amplify the efficiency and efficacy of model pre-training by concentrating dedicatedly on unmasked regions during this phase. In contrast, achieving a similar task with ConvNets poses a more intricate dilemma, necessitating the maintenance of the 2D image structure. There are two naive solutions: \textbf{(i)} the instigation of learnable masked markers at the input stage, and \textbf{(ii)} the transformation of the hierarchical convolutional network into an isotropic structure~\cite{liu2022convnet} that mirrors the ViT-style technique (\cref{tab:isotropic}). Regardless, both strategies may decrease pre-training efficiency and possibly engender inconsistencies between training and testing.

This paper is at the forefront of proposing an entirely ConvNet-based MVM approach to address this challenge. Drawing inspiration from~\cite{tian2023designing, woo2023convnext}, we utilize sparse convolutional networks to develop a masked autoencoder, which usurps traditional dense convolution. This strategy aims to meet three primary goals: \textbf{(i)} maintaining the structural integrity of the image during feature encoding, \textbf{(ii)} preventing mask dissipation in the course of hierarchical convolution, and \textbf{(iii)} markedly amplifying the computational efficiency of MVM. Specifically, our VideoMAC uses a fully convolutional architecture. Conventionally, ConvNets are structured with four stages, where the resolution of features is halved after each stage, representing a patch size of 32. For an input frame size of $224 \times 224$, the features are sequentially downsampled at every stage to achieve resolutions of $56 \times 56$, $28 \times 28$, $14 \times 14$, and $7 \times 7$, respectively. Subsequently, masks are produced and recursively upsampled to the highest resolution (\ie, $56 \times 56$) in the final stage.

Current MVMs typically necessitate significant computational resources, posing a challenge for practical applications. Our strategy, which substitutes ConvNets for ViTs, does mitigate computational complexity. Furthermore, studying existing MVM methods, we observe that to capture temporal signals within videos, most methods either utilize multi-frame data input or opt for a siamese encoder structure. We diverge from these widespread strategies and instead adopt a more resource-efficient method, namely the online-target encoder architecture (\cref{fig:pipeline}), commonly applied in self-supervised image representation learning.
The target encoder's weights are updated by the online encoder via EMA.
This architectural integration into our MVM encoding process results in three-fold advantages: \textbf{(i)} considerable reduction of computational complexity, \textbf{(ii)} joint optimization of frame-to-feature representation by the online and target encoders, and \textbf{(iii)} simplicity in incorporating existing ConvNets into our framework, thus improving the performance of downstream tasks. Consequently, visible patches $\mathcal{V}^{(1)}$ and $\mathcal{V}^{(2)}$ of consecutive frames are processed by the online encoder $f_{\mathcal{E}}^{o}$ and the target encoder $f_{\mathcal{E}}^{t}$ to generate latent representations $z^{(1)}=f_{\mathcal{E}}^{o}(\mathcal{V}^{(1)})$ and $z^{(2)}=f_{\mathcal{E}}^{t}(\mathcal{V}^{(2)})$, respectively.

\noindent\textbf{MAC Decoder.}
Unlike previous methods that utilize hierarchical decoders~\cite{tian2023designing,tian2023integrally} or transformer decoders\cite{mae_st,liu2023mixmae,tong2022videomae}, our VideoMAC implements a lightweight decoder derived from the ConvNeXt block~\cite{liu2022convnet}. This simplifies the design and bolsters the efficiency of the pre-training process. Analogous to our encoder, we construct an online decoder $f_{\mathcal{D}}^{o}$ and a target decoder $f_{\mathcal{D}}^{t}$, further employing EMA for parameter updates (\cref{fig:pipeline}). Given the latent representations $z^{(1)}$ and $z^{(2)}$ from two successive frames, decoders $f_{\mathcal{D}}^{o}$ and $f_{\mathcal{D}}^{t}$ reconstruct masked patches $\mathcal{M}^{(1)}$ and $\mathcal{M}^{(2)}$ to yield $\bm{O}^{(1)}=f_{\mathcal{D}}^{o}(z^{(1)};\mathcal{M}^{(1)})$ and $\bm{O}^{(2)}=f_{\mathcal{D}}^{t}(z^{(2)};\mathcal{M}^{(2)})$, respectively. In this paper, we employ a single-block design for the decoder, with the decoder feature projection dimension set to 512.
Detailed ablation analysis can be found in \cref{sec:ablation}, and \cref{tab:decoder_depth,tab:decoder_width}.

\begin{table*}[t]
    \centering
    \resizebox{1.0\textwidth}{!}{
    \begin{tabular}{l|cccc|ccc|c|cc}
    \hline
    \multirow{2}{*}{\begin{tabular}[c]{@{}c@{}}Method\end{tabular}} & \multirow{2}{*}{\begin{tabular}[c]{@{}c@{}}Type\end{tabular}} & \multirow{2}{*}{\begin{tabular}[c]{@{}c@{}}Backbone\end{tabular}} & \multirow{2}{*}{\begin{tabular}[c]{@{}c@{}}Training dataset\end{tabular}} & \multirow{2}{*}{\begin{tabular}[c]{@{}c@{}}Epoch\end{tabular}} & \multicolumn{3}{c|}{DAVIS17~\cite{davis17}}& VIP~\cite{zhou2018adaptive} & \multicolumn{2}{c}{JHMDB~\cite{jhuang2013towards}}  \\
    &  &  &  &  & $\mathcal{J}\&\mathcal{F}_{m}$ & $\mathcal{J}_{m}$ & $\mathcal{F}_{m}$ & mIoU & PCK@0.1 & PCK@0.2 \\
    \hline
    Supervised~\cite{he2016deep}$_{\rm{CVPR16}}$ & SL & RN50 & IN1K (~1.28M~, ~-~) & 100 & 66.0 & 63.7 & 68.4 & 39.5 & \textbf{59.2} & 78.3 \\
    \hline
    DenseCL~\cite{wang2021densecl}$_{\rm{CVPR21}}$ & CL & RN50 & IN1K (~1.28M~, ~-~) & 200 & 61.1 & 59.8 & 62.6 & 32.5 & 56.1 & 78.0 \\
    DINO~\cite{caron2021emerging}$_{\rm{ICCV21}}$ & CL & ViT-S/16 & IN1K (~1.28M~, ~-~) & 800 & 61.8 & 60.2 & 63.4 & 36.2 & 45.6 & 75.0 \\
    ODIN$^2$~\cite{henaff2022object}$_{\rm{ECCV22}}$ & CL & RN50 & IN1K (~1.28M~, ~-~) & 1000 & 54.1 & 54.3 & 53.9 & - & - & - \\
    CrOC~\cite{stegmuller2023croc}$_{\rm{CVPR23}}$ & CL & ViT-S/16 & IN1K (~1.28M~, ~-~) & 300 & 44.7 & 43.5 & 45.9 & 26.1 & - & - \\
    \hline
    MAE~\cite{he2022mae}$_{\rm{CVPR22}}$ & MIM & ViT-B/16 & IN1K (~1.28M~, ~-~) & 1600 & 53.5 & 52.1 & 55.0 & 28.1 & 44.6 & 73.4 \\
    FCMAE~\cite{woo2023convnext}$_{\rm{CVPR23}}$ & MIM & CNXv2-B & IN1K (~1.28M~, ~-~) & 1600 & 43.7 & 41.9 & 45.5 & 24.9 & 51.7 & 74.2 \\
    RC-MAE~\cite{lee2023exploring}$_{\rm{ICLR23}}$ & MIM & ViT-S/16 & IN1K (~1.28M~, ~-~) & 1600 & 49.2 & 48.9 & 50.5 & 29.7 & 43.2 & 72.3 \\
    SparK~\cite{tian2023designing}$_{\rm{ICLR23}}$ & MIM & RN50 & IN1K (~1.28M~, ~-~) & 1600 & 55.6 & 56.3 & 54.9 & 32.8 & 52.9 & 75.2 \\
    \hline
    MAE-ST~\cite{mae_st}$_{\rm{NeurIPS22}}$ & MVM & ViT-L/16 & K400 (~-~, 800 hours) & 1600 & 54.6 & 55.5 & 53.6 & 33.2 & 44.4 & 72.5 \\
    VideoMAE~\cite{tong2022videomae}$_{\rm{NeurIPS22}}$ & MVM & ViT-S/16 & K400 (~-~, 800 hours) & 1600 & 39.3 & 39.7 & 38.9 & 22.3 & 41.0 & 67.9 \\
    DropMAE~\cite{wu2023dropmae}$_{\rm{CVPR23}}$ & MVM & ViT-B/16 & K400 (~-~, 800 hours) & 1600 & 53.4 & 51.8 & 55.0 & 31.1 &42.3 & 69.2 \\
    SiamMAE~\cite{gupta2023siamese}$_{\rm{NeurIPS23}}$ & MVM & ViT-S/16 & K400 (~-~, 800 hours) & 2000 & 62.0 & 60.3 & 63.7 & 37.3 & 47.0 & 76.1 \\
    \hdashline
    \rowcolor{rowgray} \textit{Ours} &  &  &  &  &  &  &  &  &  &  \\ 
    \rowcolor{rowblue} \textbf{VideoMAC} & MVM & RN50 & YT18 (~-~, 5.58 hours) & 100 & \underline{67.2} & \underline{64.9} & \underline{69.5} & \textbf{43.6} & 57.2 & \underline{79.2} \\
    \rowcolor{rowblue} \textbf{VideoMAC} & MVM & CNXv2-S & YT18 (~-~, 5.58 hours) & 100 & \textbf{68.4} & \textbf{65.3} & \textbf{71.4}  & \underline{40.4} & \underline{58.1} & \textbf{80.0} \\
    \hline
    \end{tabular}}
    \vspace{-0.3cm}
    \leftline{~~\footnotesize{SL: Supervised Learning. CL: Contrastive Learning. IN1K: ImageNet1K~\cite{deng2009imagenet}. K400: Kinetics-400~\cite{k400}. YT18: YouTube-VOS 2018~\cite{xu2018youtube}.}}
    \caption{\textbf{Comparing VideoMAC with previous supervised and self-supervised approaches (SL, CL, MIM, and MVM)} on downstream tasks, \eg, video object segmentation on DAVIS17~\cite{davis17}, body part propagation on VIP~\cite{zhou2018adaptive}, and human pose tracking on JHMDB~\cite{jhuang2013towards}.}
    \label{tab:overall_results}
    \vspace{-0.55cm}
\end{table*}

\noindent\textbf{Reconstruction Consistency.}
In line with the majority of MAE-based methods, we compute the Mean Squared Error (MSE) between the reconstituted and original images to assess the reconstruction loss. We elect to aim for each patch of normalized pixels, executing the loss calculation solely for masked patches. As a result, we synchronize the values of normalized pixels within each image patch, applying the loss function exclusively to those patches subject to masking. Significantly, our VideoMAC yields two frames of reconstructed results from both the online and target encoders, necessitating individual loss computations:
\begin{equation}\label{eq:loss_onl_tar}
    \begin{aligned}
        \mathcal{L}_{o}=\frac{1}{|\mathcal{M}^{(1)}|}\sum_{i\in\mathcal{M}^{(1)}}||\hat{\bm{I}}_{i}^{(1)}-\bm{O}_{i}^{(1)}||_2^2, \\
        \mathcal{L}_{t}=\frac{1}{|\mathcal{M}^{(2)}|}\sum_{i\in\mathcal{M}^{(2)}}||\hat{\bm{I}}_{i}^{(2)}-\bm{O}_{i}^{(2)}||_2^2,
    \end{aligned}
\end{equation}
where $\mathcal{L}_{o}$ and $\mathcal{L}_{t}$ denote online and target losses correspondingly.
Due to the correlation between consecutive video frames, we pursue temporal consistency in addition to the reconstruction loss. Here, we introduce an intuitive assumption: if the pixel deviation between the two raw frames tends to a particular value, denoted as $\epsilon$, it is inferred that the divergence between the reconstructed frames should also trend towards $\epsilon$. Based on this assumption, we can estimate the reconstruction consistency between two frames and use it as a proxy for temporal consistency.
Hence, the reconstruction consistency loss $\mathcal{L}_{c}$ is thus expressed as:
\begin{equation}\label{eq:loss_cons}
    \begin{aligned}
        \mathcal{L}_{c}=\frac{1}{|\mathcal{M}^{(1)}|}\sum_{i\in\mathcal{M}^{(1)},~j\in\mathcal{M}^{(2)}}
||\bm{O}_{i}^{(1)}-\bm{O}_{j}^{(2)}||_2^2,
    \end{aligned}
\end{equation}
where $|\mathcal{M}^{(1)}|$ can be substituted by $|\mathcal{M}^{(2)}|$, courtesy of our utilization of symmetric masking. Ultimately, our approach culminates in a total loss expressed as:
\begin{equation}\label{eq:loss_total}
    \begin{aligned}
        \mathcal{L}_{total}=\mathcal{L}_{o} + \mathcal{L}_{t} + \gamma\mathcal{L}_{c},
    \end{aligned}
\end{equation}
with $\gamma$ symbolizing the weight factor of the consistency loss (see \cref{tab:rec_cons} and \cref{fig:cons_loss} for detailed analysis).
Accordingly, our approach, VideoMAC, accomplishes the realization of a fully convolutional network MVM. Moreover, the reconstruction consistency loss introduced in this paper provides a promising solution for modeling spatio-temporal data.

\subsection{Implementation}
\noindent\textbf{Architecture.}
Our VideoMAC possesses the adaptability to employ any convolutional network the an encoder. To permit a fair comparison, we choose two representative convolutional networks for our experiments: ResNet (RN)~\cite{he2016deep} and ConvNeXtv2 (CNXv2)~\cite{woo2023convnext}. Although RN50 (26M) and CNXv2-S (50M) serve as the primary experimental models, we further investigate models of diverse sizes (\cref{tab:isotropic,tab:encoder_type,tab:finetune_in1k}). In this paper, the decoder component involves a readily available ConvNeXt block~\cite{liu2022convnet}, thereby dispensing with the necessity for intricate network design. During pre-training, we employ an encoder implemented with the MinkowskiEngine sparse convolution library~\cite{choy20194d}. It's noteworthy that the sparse convolution layer can be seamlessly converted back to a standard convolution layer in downstream tasks without requiring additional processes.

\begin{table}[t]
    \vspace{0.1cm}
    \centering
    \setlength{\tabcolsep}{4.3pt}
    \resizebox{1.0\linewidth}{!}{
    \begin{tabular}{lcccc}
    \hline
    MVM Method & Resolution & \#Param. & FLOPs & $\mathcal{J}\&\mathcal{F}_{m}$ \\
    \hline
    \rowcolor{rowgray} \textit{Iso. ViT Backbone} & & & & \\
    MAE-ST~\cite{mae_st} & $14\times14$ & 304M & 61.6G & 54.6 \\
    VideoMAE~\cite{tong2022videomae} & $14\times14$ & \underline{22M} & 4.6G& 39.3 \\
    DropMAE~\cite{wu2023dropmae} & $14\times14$ & 87M & 17.6G & 53.4 \\
    SiamMAE~\cite{gupta2023siamese} & $14\times14$ & \underline{22M} & 4.6G & 62.0 \\
    \hline
    \rowcolor{rowgray} \textit{Iso. ConvNet Backbone} & & & & \\
    \textbf{VideoMAC} (CNXv2-S$^\dag$) & $14\times14$ & \underline{22M} & 4.3G & 53.8 \\
    \hline
    \rowcolor{rowgray} \textit{Hie. ConvNet Backbone} & & & & \\
    \rowcolor{rowblue} \textbf{VideoMAC} (RN18) & $7\times7$ & \textbf{12M} & \textbf{1.8G} & 64.7 \\
    \rowcolor{rowblue} \textbf{VideoMAC} (RN50) & $7\times7$ & 26M & \underline{4.1G} & 67.2 \\
    \rowcolor{rowblue} \textbf{VideoMAC} (CNXv2-T) & $7\times7$ & 29M & 4.5G & \underline{67.5} \\
    \rowcolor{rowblue} \textbf{VideoMAC} (CNXv2-S) & $7\times7$ & 50M & 8.7G & \textbf{68.4} \\
    \hline
    \end{tabular}}
    \vspace{-0.3cm}
    \caption{\textbf{Comparisons with isotropic (\textit{iso.}) and hierarchical (\textit{hie.})} based MVM methods. `\dag' denotes our modified \textit{ios.} version.}
    \label{tab:isotropic}
    \vspace{-0.6cm}
\end{table}

\noindent\textbf{Pre-training.}
Different from previous methods that pre-train on the large-scale video dataset K400~\cite{k400}, VideoMAC pre-trains on YouTube-VOS~\cite{xu2018youtube} for 100 epochs, leading to a significant reduction in pre-training time by over 140 times. During VideoMAC pre-training, the standard input comprises two frames with a spatial size of $224\times224$ pixels. The frame interval for this study is determined to be 1 (\cref{tab:frame_gap}), implying consecutive frames are chosen. We employ minimal data augmentation, encompassing only random resized cropping and horizontal flipping. A masking ratio of 0.75 is utilized to pre-train our models.
We conduct training using an AdamW optimizer~\cite{loshchilov2017adamw} with a batch size of 512 and a learning rate of 2e-3. Cosine decay~\cite{loshchilov2016sgdr} is applied to adjust the learning rate.

\noindent\textbf{Downstream Tasks.}
To gauge the efficacy of VideoMAC at video representation learning, we undertake comprehensive experiments on three video downstream tasks, namely video object segmentation, body part propagation, and human pose tracking (\cref{tab:overall_results}).
We also investigate the performance of image recognition through video pre-training. Specifically, our backbone is pre-trained on video data and then fine-tuned on ImageNet1K~\cite{deng2009imagenet} (\cref{tab:finetune_in1k}).

\begin{figure*}
    \begin{center}
        \includegraphics[width=1.0\linewidth]{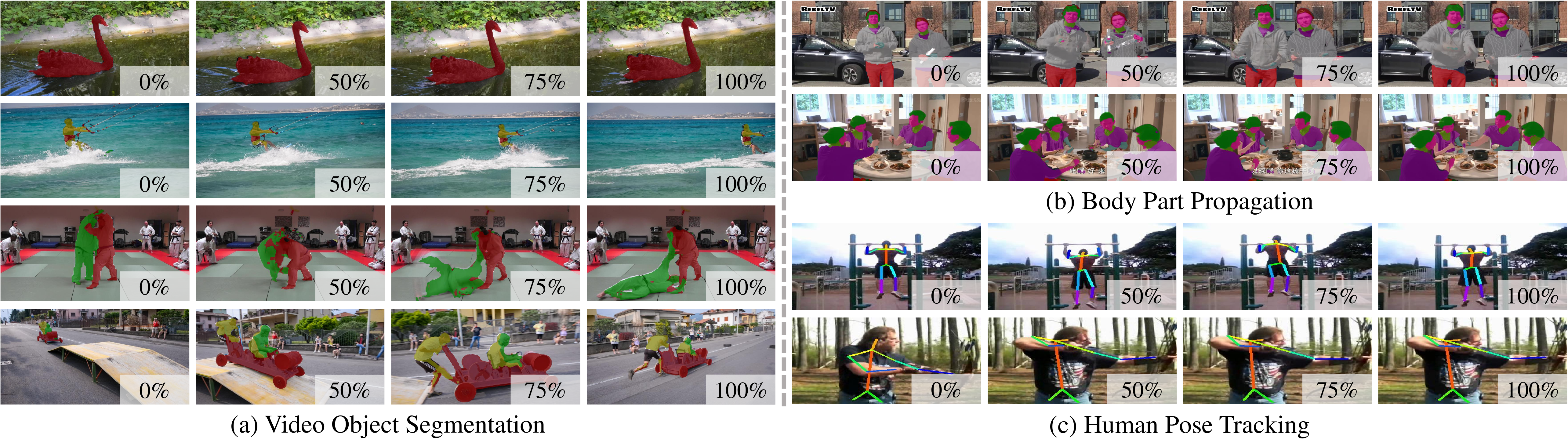}
    \end{center}
    \vspace{-0.7cm}
    \caption{\textbf{Qualitative results of our VideoMAC} (using CNXv2-S) for three video downstream tasks: (a) video object segmentation on DAVIS17~\cite{davis17}), (b) body part propagation on VIP~\cite{zhou2018adaptive}, and (c) human pose tracking on JHMDB~\cite{jhuang2013towards}.}\label{fig:result_vis}
    \vspace{-0.2cm}
\end{figure*}

\begin{table*}[t]
  \centering
  \begin{minipage}{0.64\textwidth}
    \centering
    \subfloat[\textbf{Encoder type}. VideoMAC is compatible with both classical and modern ConvNets. \label{tab:encoder_type}]{
        \begin{minipage}{0.45\linewidth}
            \centering
            \setlength{\tabcolsep}{5.5pt}
            \resizebox{1.0\textwidth}{!}{
            \begin{tabular}{lccc}
                \hline
                Backbone & $\mathcal{J}\&\mathcal{F}_{m}$ & $\mathcal{J}_{m}$ & $\mathcal{F}_{m}$ \\
                \hline
                R18 & 64.7 & 62.8 & 66.5 \\
                R50 & 67.2 & 64.9 & 69.5 \\
                CNXv2-T & 67.5 & 64.4 & 70.6 \\
                \rowcolor{rowblue} CNXv2-S & \textbf{68.4} & \textbf{65.3} & \textbf{71.4} \\
                \hline
            \end{tabular}}
        \end{minipage}
    }
    \hspace{0.4cm}
    \subfloat[\textbf{Decoder depth}. One block achieves competitive performance and efficiency. \label{tab:decoder_depth}]{
        \begin{minipage}{0.45\linewidth}
            \centering
            \setlength{\tabcolsep}{7.2pt}
            \resizebox{1.0\textwidth}{!}{
            \begin{tabular}{cccc}
                \hline
                Blocks & $\mathcal{J}\&\mathcal{F}_{m}$ & $\mathcal{J}_{m}$ & $\mathcal{F}_{m}$ \\
                \hline
                \rowcolor{rowblue} 1 & \textbf{68.4} & 65.3 & \textbf{71.4} \\
                2 & 68.2 & 65.2 & 71.2 \\
                4 & 68.3 & \textbf{65.4} & 71.2 \\
                8 & 68.1 & 65.2 & 71.0 \\
                \hline
            \end{tabular}}
        \end{minipage}
    }
    \\
    \subfloat[\textbf{Decoder width}. A decoder width of 512 delivers the best performance. \label{tab:decoder_width}]{
        \begin{minipage}{0.45\linewidth}
            \centering
            \setlength{\tabcolsep}{8pt}
            \resizebox{1.0\textwidth}{!}{
            \begin{tabular}{cccc}
                \hline
                Dim. & $\mathcal{J}\&\mathcal{F}_{m}$ & $\mathcal{J}_{m}$ & $\mathcal{F}_{m}$ \\
                \hline
                128 & 67.9 & 64.8 & 71.0 \\
                256 & 68.2 & \textbf{65.3} & 71.1 \\
                \rowcolor{rowblue} 512 & \textbf{68.4} & \textbf{65.3} & \textbf{71.4} \\
                1024 & 68.1 & 65.0 & 71.2 \\
                \hline
            \end{tabular}}
        \end{minipage}
    }
    \hspace{0.4cm}
    \subfloat[\textbf{Masking design}. Symmetric (sym.) masking outperforms asymmetric (asym.) one. \label{tab:asym_mask}]{
        \begin{minipage}{0.45\linewidth}
            \centering
            \setlength{\tabcolsep}{6pt}
            \resizebox{1.0\textwidth}{!}{
            \begin{tabular}{cccc}
                \hline
                Masking & $\mathcal{J}\&\mathcal{F}_{m}$ & $\mathcal{J}_{m}$ & $\mathcal{F}_{m}$ \\
                \hline
                Asym. & 66.5 & 63.4 & 69.6 \\
                \rowcolor{rowblue} Sym. & \textbf{68.4} & \textbf{65.3} & \textbf{71.4} \\
                \hline
                & & & \\
                & & & \\
            \end{tabular}}
        \end{minipage}
    }
    \vspace{-0.3cm}
    \caption{\textbf{VideoMAC encoder and decoder ablation experiments} on DAVIS17~\cite{davis17}. We report ablation results for two of the most emblematic families of ConvNets. Unless specified otherwise, the encoder is CNXv2-S in default. `Dim' indicates the number of decoder dimensions. Default settings employed in this paper are marked in \colorbox{rowblue}{blue}.} \label{tab:coder_ablations}
    \end{minipage}
    \hspace{0.4cm}
    \begin{minipage}{0.32\textwidth}
    \centering
    \includegraphics[width=1.0\textwidth]{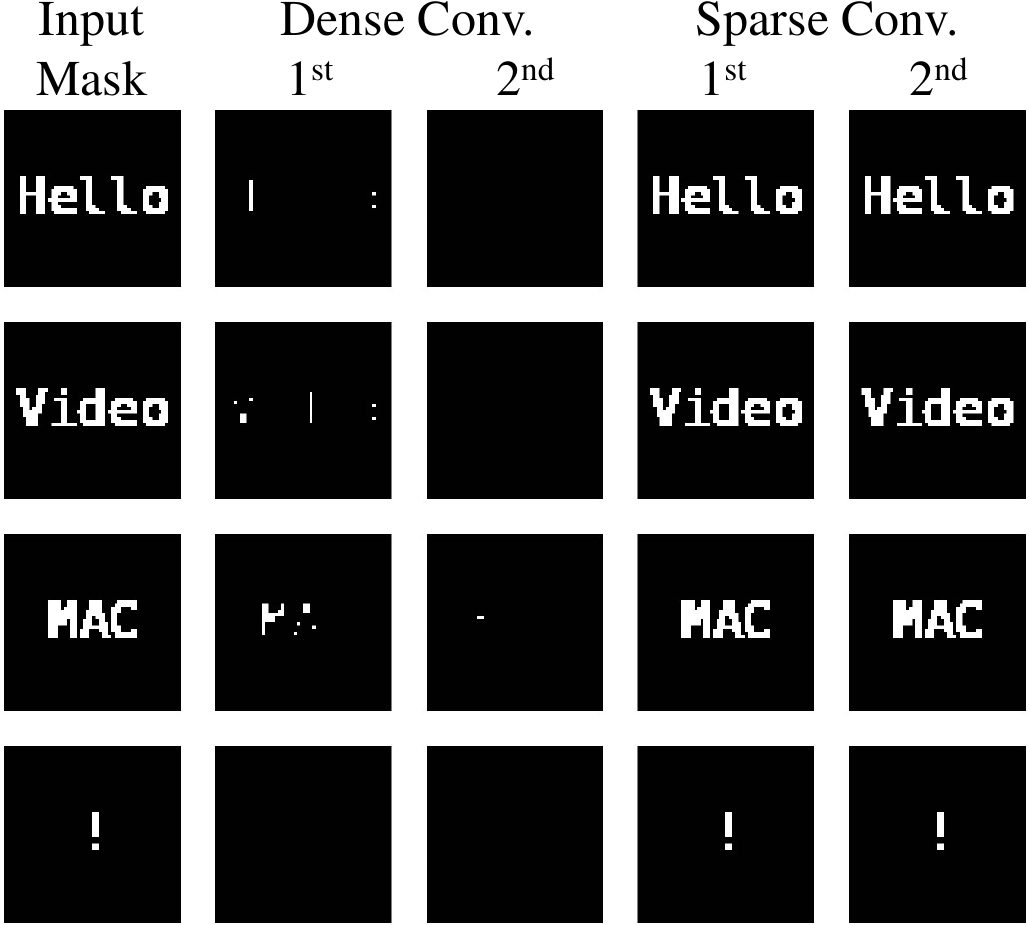}
    \vspace{-0.43cm}
    \captionof{figure}{For masked modeling, dense convolution usually results in the dissipation of mask structures. The deployment of sparse convolution proves to be an effective solution, enabling ConvNet-based MIM / MVM.}
    \label{fig:sparse_conv}
    \end{minipage}
    \vspace{-0.47cm}
\end{table*}

\section{Experiments}
This section begins with assessing VideoMAC's performance in three downstream tasks, comparing it to previous state-of-the-art methods. Subsequently, we conduct extensive ablation studies to elucidate the effectiveness of our proposed approach. The final part of this section explores VideoMAC's ability for image recognition, accompanied by a visualization of the reconstruction results.

\subsection{Comparison with Previous methods} 
\noindent\textbf{Video Object Segmentation.}
We evaluate VideoMAC on the popular DAVIS17~\cite{davis17} dataset, which includes a validation set comprising 30 videos. Our evaluation methodology aligns with \cite{jabri2020space,caron2021emerging}, and we report the typical evaluation metrics~\cite{davis16}, including the mean region similarity $\mathcal{J}_{m}$, the mean boundary accuracy $\mathcal{F}_{m}$, and their average values $\mathcal{J}\&\mathcal{F}_{m}$. As shown in \cref{tab:overall_results}, our VideoMAC achieves state-of-the-art results for video object segmentation. With a smaller video training dataset and fewer training epochs, both our VideoMAC models (\ie, CNXv2-S and RN50) surpass the SiamMAE~\cite{gupta2023siamese} by \textbf{6.4\%} and \textbf{5.2\%}, respectively, in terms of $\mathcal{J}\&\mathcal{F}_{m}$. It's worth noting that our approach achieves superior performance to supervised method~\cite{he2016deep} based on MVM, with improvements of \textbf{2.4\%} and \textbf{1.4\%}, respectively. Owing to the hierarchical feature extraction capabilities inherent in ConvNets, our approach demonstrates superior proficiency in spatial information extraction in contrast to ViT-based MVM methods~\cite{mae_st,tong2022videomae,wu2023dropmae,gupta2023siamese}. Concurrently, the introduced reconstruction consistency framework strengthens the acquisition of temporal correlations.

\noindent\textbf{Body Part Propagation.}
We evaluate VideoMAC on the Video Instance Parsing (VIP) dataset~\cite{zhou2018adaptive}, and the quantitative results, measured by mIoU, are detailed in \cref{tab:overall_results}. This validation set comprises of 50 videos that concentrate on the propagation of 19 body parts, such as arms and legs, thereby necessitating a heightened level of precision in matching as compared to video object segmentation. We adhere to the same parameters as \cite{jabri2020space}, adjusting the video frames to $560 \times 560$ dimensions. In the case of body part propagation, two variants of our VideoMAC demonstrate improvements by \textbf{3.1\%} (CNXv2-S) and \textbf{6.3\%} (RN50), respectively, in comparison with the top-performing SiamMAE~\cite{gupta2023siamese} in MVM.
Significantly, our RN50-based model outperforms CNXv2-S on the VIP dataset~\cite{zhou2018adaptive} due to the higher dimensionality of RN50 features (2048 vs. 768), which benefits complex tasks and improves accuracy in mask propagation.

\noindent\textbf{Human Pose Tracking.}
We conduct the human pose tracking task on the validation set of JHMDB~\cite{jhuang2013towards}, which includes 268 videos and entails detecting 15 human keypoints. Conforming to the protocol delineated in \cite{jabri2020space}, we resize video frames to a resolution of $320 \times 320$, and employ the Possibility of Correct Keypoints (PCK) metric \cite{song2017thin} for quantitative evaluation. As shown in \cref{tab:overall_results}, both variants (\ie, RN50 and CNXv2-S) of our VideoMAC outperforms all MVM methods in terms of PCK@0.1 and PCK@0.2.

\noindent\textbf{Qualitative results.}
As depicted in \cref{fig:result_vis}, we present the qualitative results of mask propagation achieved by VideoMAC on three downstream tasks. Our VideoMAC yields notable performance across a spectrum of complex videos.

\begin{table*}[t!]
    \centering
    \subfloat[\textbf{Training Dataset}. Our VideoMAC pre-trained on YT18 yields better performance and efficiency compared to K400. \label{tab:training_data}]{
        \begin{minipage}{0.39\linewidth}
            \centering
            \resizebox{1.0\textwidth}{!}{
            \begin{tabular}{lcccc}
                \hline
                Dataset & Size & $\mathcal{J}\&\mathcal{F}_{m}$ & $\mathcal{J}_{m}$ & $\mathcal{F}_{m}$ \\
                \hline
                \rowcolor{rowblue} YT18~\cite{xu2018youtube} & 5.58 hours & \textbf{68.4} & \textbf{65.3} & \textbf{71.4} \\
                K400\textcolor{red}{\footnotemark[1]}~\cite{k400} & 800 hours & 67.8 & 65.2 & 70.3 \\
                \hline
                & & & & \\
            \end{tabular}}
        \end{minipage}
    }
    \hspace{0.1cm}
    \subfloat[\textbf{Data augmentation}. Minimal data augmentation achieves the best performance. \label{tab:data_aug}]{
        \begin{minipage}{0.315\linewidth}
            \centering
            \resizebox{1.0\textwidth}{!}{
            \begin{tabular}{lccc}
                \hline
                Augmentation & $\mathcal{J}\&\mathcal{F}_{m}$ & $\mathcal{J}_{m}$ & $\mathcal{F}_{m}$ \\
                \hline
                \rowcolor{rowblue} spatial & \textbf{68.4} & \textbf{65.3} & \textbf{71.4} \\
                color & 64.9 & 61.8 & 68.0 \\
                spatial + color & 66.1 & 62.9 & 69.3 \\
                \hline
            \end{tabular}}
        \end{minipage}
    }
    \hspace{0.1cm}
    \subfloat[\textbf{Frame gap}. Video pairs with one frame gap perform the best results. \label{tab:frame_gap}]{
        \begin{minipage}{0.243\linewidth}
            \centering
            \resizebox{1.0\textwidth}{!}{
            \begin{tabular}{lccc}
                \hline
                Gap & $\mathcal{J}\&\mathcal{F}_{m}$ & $\mathcal{J}_{m}$ & $\mathcal{F}_{m}$ \\
                \hline
                \rowcolor{rowblue} 1 & \textbf{68.4} & \textbf{65.3} & \textbf{71.4} \\
                5 & 66.4 & 63.5 & 69.2 \\
                10 & 63.2 & 60.4 & 65.9 \\
                \hline
            \end{tabular}}
        \end{minipage}
    }
    \\
    \vspace{0.08cm}
    \subfloat[\textbf{Model loss}. VideoMAC utilizes consistency loss $\mathcal{L}_{c}$ and target loss $\mathcal{L}_{t}$ for the best. \label{tab:target_loss}]{
        \begin{minipage}{0.30\linewidth}
            \centering
            \setlength{\tabcolsep}{7.5pt}
            \resizebox{1.0\textwidth}{!}{
            \begin{tabular}{ccccc}
                \hline
                $\mathcal{L}_{c}$ & $\mathcal{L}_{t}$ & $\mathcal{J}\&\mathcal{F}_{m}$ & $\mathcal{J}_{m}$ & $\mathcal{F}_{m}$ \\
                \hline
                \ding{55} & \ding{55} & 51.1 & 52.2 & 50.0 \\
                \ding{55} & \ding{51} & 60.6 & 56.3 & 64.9 \\
                \ding{51} & \ding{55} & 63.4 & 63.4 & 68.7 \\
                \rowcolor{rowblue} \ding{51} & \ding{51} & \textbf{68.4} & \textbf{65.3} & \textbf{71.4} \\
                \hline
            \end{tabular}}
        \end{minipage}
    }
    \hspace{0.3cm}
    \subfloat[\textbf{Reconstruction consistency}. When $\gamma$ is set to 1.0, the consistency loss works the best.\label{tab:rec_cons}]{
        \begin{minipage}{0.30\linewidth}
            \centering
            \setlength{\tabcolsep}{5.8pt}
            \resizebox{1.0\textwidth}{!}{
            \begin{tabular}{cccc}
                \hline
                Consistency & $\mathcal{J}\&\mathcal{F}_{m}$ & $\mathcal{J}_{m}$ & $\mathcal{F}_{m}$ \\
                \hline
                $\gamma$=0.0 & 60.6 & 56.3 & 64.9 \\
                $\gamma$=0.1 & 64.8 & 62.0 & 67.6 \\
                $\gamma$=0.5 & 65.8 & 63.2 & 68.5 \\
                \rowcolor{rowblue} $\gamma$=1.0 & \textbf{68.4} & \textbf{65.3} & \textbf{71.4} \\
                \hline
            \end{tabular}}
        \end{minipage}
    }
    \hspace{0.3cm}
    \subfloat[\textbf{Masking ratio}. Our VideoMAC uses a masking ratio of 0.75 for the highest performance. \label{tab:mask_ratio}]{
        \begin{minipage}{0.32\linewidth}
            \centering
            \setlength{\tabcolsep}{6.3pt}
            \resizebox{1.0\textwidth}{!}{
            \begin{tabular}{cccc}
                \hline
                Masking ratio & $\mathcal{J}\&\mathcal{F}_{m}$ & $\mathcal{J}_{m}$ & $\mathcal{F}_{m}$ \\
                \hline
                0.65 & 66.5 & 63.8 & 69.1 \\
                \rowcolor{rowblue} 0.75 & \textbf{68.4} & \textbf{65.3} & \textbf{71.4} \\
                0.85 & 68.3 & \textbf{65.3} & 71.3 \\
                0.95 & 60.3 & 59.7 & 60.9 \\
                \hline
            \end{tabular}}
        \end{minipage}
    }
    \vspace{-0.3cm}
    \caption{\textbf{VideoMAC ablations} on data, loss, and masking ratio configuration. We report two official metrics $\mathcal{J}_{m}$ and $\mathcal{F}_{m}$ on DAVIS17~\cite{davis17}.}
    \label{tab:data_loss_ratio}
    \vspace{-0.4cm}
\end{table*}

\subsection{Ablation Studies} \label{sec:ablation}
In this section, we conduct extensive ablation studies to analyze the component design of our VideoMAC.

\noindent\textbf{Encoder design: \textit{hierarchical} vs. \textit{isotropic}.}
In this ablation study, we scrutinize diverse encoder architectures to our VideoMAC, \ie, hierarchical and isotropic designs. Precisely, we adopted the ViT~\cite{dosovitskiy2021vit} design principles and restructured the CNXv2-S backbone from having a hierarchical structure into an isotropic one. This implies the creation of an architecture devoid of downsampling layers, ensuring consistent feature resolution (\eg, $14 \times 14$) throughout all depths. As shown in \cref{tab:isotropic}, the performance difference between our VideoMAC (\textit{iso.}) and the ViT-based MVM methods (\eg, VideoMAE~\cite{tong2022videomae} and DropMAE~\cite{wu2023dropmae}) is marginal. By maintaining the hierarchical structure of ConvNets within VideoMAC models (\textit{hie.}), we improve the performance significantly, even though their feature resolution (\ie, $7 \times 7$) is merely half that of the ViT-based models~\cite{mae_st,tong2022videomae,wu2023dropmae,gupta2023siamese}. This can be ascribed to ConvNets' capacity to preserve multi-scale spatial information.

\noindent\textbf{Encoder backbone: \textit{classical} and \textit{modern}.}
For our study, we incorporate two emblematic ConvNet families, \ie, ResNets~\cite{he2016deep} and ConvNeXts~\cite{liu2022convnet,woo2023convnext}. As shown in \cref{tab:encoder_type}, we substitute dense convolution with sparse convolution in the baseline ConvNets. This adaptation seamlessly dovetails into our VideoMAC framework, thereby facilitating a ConvNet-based MVM. In addition, \cref{fig:sparse_conv} highlights the influence of employing dense and sparse convolution on masks. It is apparent that sparse convolution effectively safeguards the structural integrity of the mask, thus enabling the integration of MAE pre-training within ConvNets.

\begin{figure}
    \begin{center}
        \includegraphics[width=1.0\linewidth]{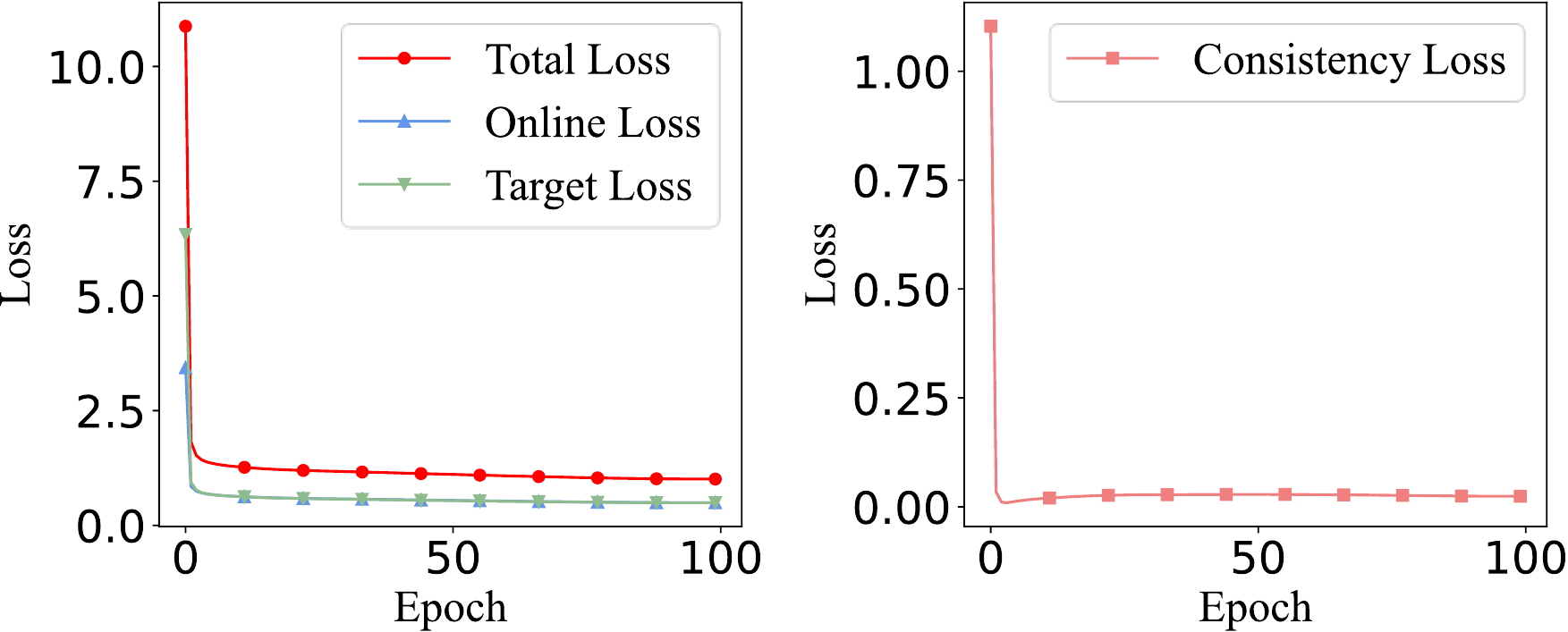}
    \end{center}
    \vspace{-0.7cm}
    \caption{\textbf{Analysis of individual loss components} after VideoMAC pre-training 100 epochs on YouTube-VOS 2018~\cite{xu2018youtube}.}
    \label{fig:loss}
    \vspace{-0.665cm}
\end{figure}

\noindent\textbf{Decoder design: \textit{blocks} and \textit{dimensions}.} 
Adhering to a simplistic design principle, our VideoMAC adopts the decoder from the purely convolutional network structure deployed in ConvNeXt~\cite{liu2022convnet}. We conduct ablation experiments focusing on the depth and width of the decoder, with the results detailed in \cref{tab:decoder_depth,tab:decoder_width}. The decoder exhibits competitive results with a single block and 512 dimensions, striking a trade-off between performance and efficiency.

\noindent\textbf{Masking design: \textit{asymmetric} vs. \textit{symmetric}.} 
In \cref{tab:asym_mask}, we investigate the impact of two masking strategies, namely asymmetric (same masking ratio but different locations) and symmetric (same masking ratio and locations), on performance. Compared to asymmetric masking, symmetric masking simplifies the calculation of the reconstruction loss, as it covers the entire masked area. Moreover, it facilitates the inclusion of more reconstruction regions in the loss calculation, resulting in faster model convergence.

\begin{figure}
    \begin{center}
        \includegraphics[width=1.0\linewidth]{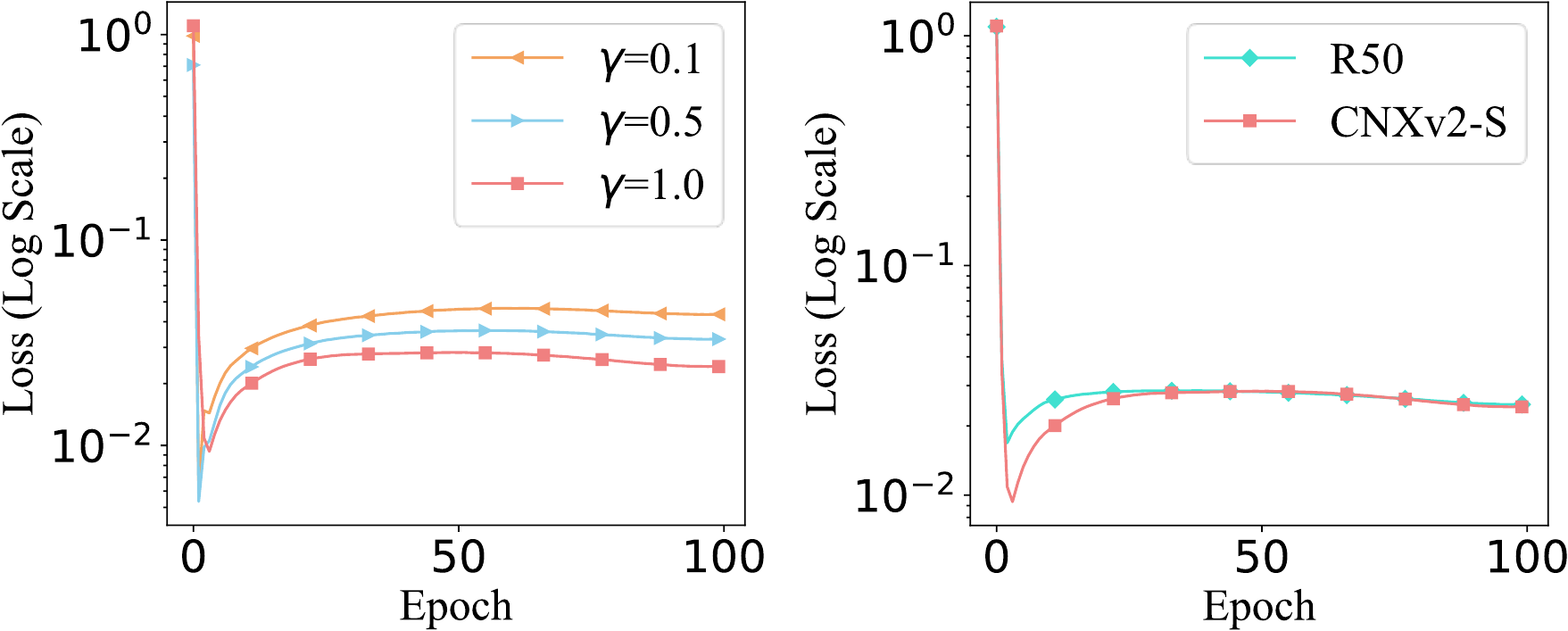}
    \end{center}
    \vspace{-0.7cm}
    \caption{\textbf{Consistency loss} in relation to $\gamma$ (left: identical encoder, CNXv2-S) and encoders (right: identical weight factor, $\gamma=1.0$).}
    \label{fig:cons_loss}
    \vspace{-0.6cm}
\end{figure}

\noindent\textbf{Data setting: \textit{dataset}, \textit{augmentation}, and \textit{gap}.}
To assess the influence of different video datasets, \cref{tab:training_data} presents the results of pre-training our VideoMAC on both YT18~\cite{xu2018youtube} and K400\footnote{K400 is sampled to match YT18's training data volume.}~\cite{k400}. 
The ablation results of VideoMAC on the two datasets differ by just \textbf{0.6\%}, demonstrating the robustness and adaptability of our method to different datasets.
For data augmentation, as shown in \cref{tab:data_aug}, our VideoMAC achieves the best results using only spatial strategy, \ie, random resized cropping and horizontal flipping.
Additionally, we conduct an ablation study on the frame gap, and the optimal results are obtained with consecutive frames (\ie, the gap is 1), as indicated in \cref{tab:frame_gap}.

\begin{figure*}
    \begin{center}
        \includegraphics[width=1.0\linewidth]{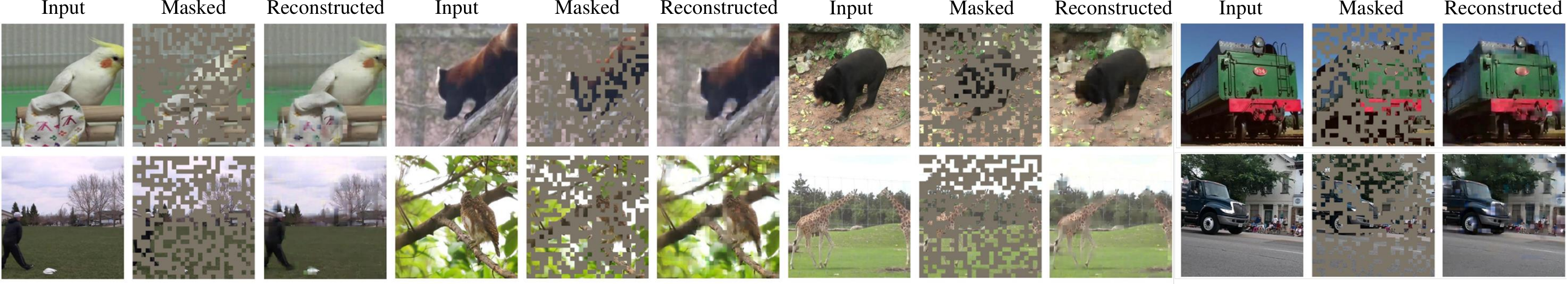}
    \end{center}
    \vspace{-0.7cm}
    \caption{\textbf{Visualization of reconstruction examples} using a pre-trained VideoMAC (CNXv2-S) model with 0.75 masking ratio. These frames are randomly selected from the validation set of YT18~\cite{xu2018youtube} and masked 60\% of patches.}\label{fig:rec_viz}
    \vspace{-0.5cm}
\end{figure*}

\noindent\textbf{The contribution of $\mathcal{L}_{c}$ and $\mathcal{L}_{t}$.}
Previous MVM methods predominantly rely on a single loss, \ie, the reconstruction loss, necessitating multiple frames to be simultaneously processed by the model. 
In contrast, our study utilizes a more computation-friendly strategy by inputting frame pairs into two separate encoders, followed by parameter updates using EMA.
Consequently, we introduce two distinct loss terms: online $\mathcal{L}_{o}$ and target $\mathcal{L}_{t}$ reconstruction losses. Furthermore, to integrate temporal information, we compute the loss between the reconstructed results of frame pairs, colloquially known as the reconstruction consistency loss $\mathcal{L}_{c}$. As shown in \cref{tab:target_loss}, $\mathcal{L}_{t}$ and $\mathcal{L}_{c}$ improve \textbf{9.5\%} and \textbf{12.3\%} in terms of $\mathcal{J}\&\mathcal{F}_{m}$, respectively, with the full VideoMAC model yielding the best results. The diverse loss components during pre-training are illustrated in \cref{fig:loss}.

\noindent\textbf{The effect of weight factor $\gamma$.} 
To investigate the impact of the weight factor on reconstruction consistency loss $\mathcal{L}_{c}$, we conduct experiments with different factor settings, \ie, $\gamma\in\{0.0, 0.1, 0.5, 1.0\}$. As illustrated in \cref{tab:rec_cons}, superior performance is attained when $\gamma$ approximates to 1. When $\gamma$ equals 0, which signifies the absence of the reconstruction consistency loss term, a detrimental effect on performance is noticeable. Simultaneously, we record $\mathcal{L}_{c}$ for various weight factors during pre-training, as depicted in \cref{fig:cons_loss}. It becomes evident that an elevated value of $\gamma$ results in a lesser loss. Ultimately, both the R50 and CNXv2-S models converge to comparable final losses. It is noteworthy that $\mathcal{L}_{c}$ presents an initial decrease, succeeded by an increase, and ultimately reaches convergence, which aligns with the description in \cref{eq:loss_cons}, implying that the ultimate reconstruction results between two frames converge to $\epsilon$.

\noindent\textbf{The impact of masking ratio.}
An ablation analysis of the masking ratio is conducted, and the results for our VideoMAC models (CNXv2-S) are documented in \cref{tab:mask_ratio}.
Ablation results portray a performance pattern that first ascends and subsequently descends with increasing masking ratios. Noticeably, the performance of our approach experiences a drastic decline at a 0.95 masking ratio caused by overfitting to the self-supervised pretext task.

\subsection{Additional Experiments} \label{sec:add_exp}
\noindent\textbf{VideoMAC for Image Recognition.}
As shown in \cref{tab:finetune_in1k}, utilizing CNXv1-T~\cite{liu2022convnet} / CNXv2-T~\cite{woo2023convnext} pre-trained by our VideoMAC on YT18~\cite{xu2018youtube}, we achieve \textbf{82.3\%} / \textbf{82.9\%} accuracy on IN1K~\cite{deng2009imagenet} image classification. In comparison to CNXv1-T and CNXv2-T trained from scratch, our approach improves the image recognition results by \textbf{0.2\%} and\textbf{ 0.4\%}. Remarkably, the performance garnered by VideoMAC (MVM) is proximate to or commensurate with the state-of-the-art ConvNet-based MIM methods~\cite{tian2023designing,woo2023convnext}. Given the pre-training cost (100 epochs vs. 1600 epochs) and the domain gap (video vs. image), we regard this improvement of our VideoMAC as noteworthy and promising compared to training from scratch and pre-training by MIM.

\begin{table}[t]
    \vspace{0.2cm}
    \centering
    \resizebox{1.0\linewidth}{!}{
    \begin{tabular}{lcccc}
    \hline
    Method & Backbone & PT epoch & PT data & FT acc. \\
    \hline
    Supervised~\cite{liu2022convnet} & CNXv1-T & - & - & 82.1 \\
    FCMAE~\cite{woo2023convnext} & CNXv1-T & 1600 & IN1K & 82.3 (+0.2) \\
    Spark~\cite{tian2023designing} & CNXv1-T & 1600 & IN1K & 82.4 (+0.3) \\
    \rowcolor{rowblue} VideoMAC & CNXv1-T & 100 & YT18 & 82.3 (\textcolor{highgreen}{+0.2}) \\
    \hline
    Supervised~\cite{woo2023convnext} & CNXv2-T & - & - & 82.5 \\
    FCMAE~\cite{woo2023convnext} & CNXv2-T & 1600 & IN1K & 83.0 (+0.5) \\
    \rowcolor{rowblue} VideoMAC & CNXv2-T & 100 & YT18 & 82.9 (\textcolor{highgreen}{+0.4}) \\
    \hline
    \end{tabular}}
    \vspace{-0.3cm}
    \caption{\textbf{Comparing VideoMAC with previous ConvNet-based MIM approaches} (\eg, Spark~\cite{tian2023designing} and FCMAE~\cite{woo2023convnext}) on the image downstream task, \ie, fine-tuning recognition accuracy (acc.) on IN1K~\cite{deng2009imagenet}. `PT' and `FT' indicate pre-training and fine-tuning.}
    \label{tab:finetune_in1k}
    \vspace{-0.6cm}
\end{table}

\noindent\textbf{Visualization.}
We visualize reconstruction results from the YT18~\cite{xu2018youtube} validation set to analyze performance in MVM pre-training. In \cref{fig:rec_viz}, VideoMAC makes reasonable predictions about the masked regions, \eg, color and contour.

\section{Conclusion}
In this study, we present VideoMAC, an MVM pre-training framework constructed fully using ConvNets. Our masked modeling for frame pairs is executed by introducing an online network (encoder and decoder) optimized by gradients, as well as the target network that updates parameters via EMA. Concurrently, for reconstruction results derived from both online and target networks, our proposed loss of reconstruction consistency between frame pairs paves the way for temporal modeling. Our VideoMAC empowers a multitude of ConvNets to reap the performance gains for downstream tasks (\eg, video but also image) by MVM pre-training.

\noindent\textbf{Limitation and Future Work.}
Existing ConvNet-based architectures typically use a fixed downsampling ratio, which in turn leads to a relatively fixed patch size. For future work, we will investigate ConvNets with adjustable patch size, allowing for higher feature resolution like ViT.

\small{\noindent\textbf{Acknowledgement.}
This work was supported by the National Natural Science Foundation of China (No. 62102182, 62202227, 62302217), Natural Science Foundation of Jiangsu Province (No. BK20220934, BK20220938, BK20220936), China Postdoctoral Science Foundation (No. 2022M721626, 2022M711635), Jiangsu Funding Program for Excellent Postdoctoral Talent (No. 2022ZB267), Fundamental Research Funds for the Central Universities (No.30923010303).}

{
    \small
    \bibliographystyle{ieeenat_fullname}
    \bibliography{main}
}


\end{document}